# வேர்ச்சொல் (VerChol)

## Grammar-First Tokenization for Agglutinative Languages

*A Morphological Alternative to Statistical Subword Tokenization, with Full Tamil Wikipedia Evaluation*

**Prabhu Raja**

Bengaluru, India

# Abstract

Tokenization is the foundational step in all large language model (LLM) pipelines, yet the dominant approach—Byte-Pair Encoding (BPE) and its variants—is inherently script-agnostic and optimized for English-like morphology. For *agglutinative* languages—a typological class encompassing the Dravidian family (Tamil, Kannada, Telugu, Malayalam), Turkic languages (Turkish, Azerbaijani, Uzbek), Uralic languages (Finnish, Hungarian, Estonian), Korean, Japanese, Swahili, Basque, and others—a single word may encode root, tense, aspect, person-number-gender agreement, case, and postpositions into one orthographic unit. Statistical tokenizers fragment these words into byte-pair chunks that sever morpheme boundaries and inflate token counts.

This paper introduces **VerChol** (வேர்ச்சொல், literally "root-word" in Tamil), a language-parametric architecture for morphological tokenization designed from linguistic first principles. VerChol employs a four-tier pipeline: (1) whole-word vocabulary lookup for known inflected forms, (2) rule-based morphological decomposition into roots and suffixes, (3) phonologically-valid syllable segmentation, and (4) character-level fallback. Unlike BPE, every token produced by VerChol is a linguistically meaningful unit—a morpheme, syllable, or character—never an arbitrary byte-sequence fragment.

We validate VerChol on Tamil as a representative agglutinative language, using the full Tamil Wikipedia corpus (774 MB, 30.5 million word occurrences, 1.85 million unique word types). On 483,313 unique words with frequency ≥ 3, VerChol with a 32,991-token vocabulary achieves a **fertility of 1.86 tokens per word**, compared to 2.85 for SentencePiece BPE (16K vocab) and 3.52 for a production Indic-optimized BPE tokenizer (68K vocab). This represents **35% fewer tokens than standard BPE and 47% fewer than the Indic-optimized baseline**, achieved with zero training compute—the entire vocabulary is constructed from a linguistic dictionary and morphological rule system. The architecture is language-parametric: the pipeline logic is invariant, while language-specific components (root dictionary, suffix catalog, sandhi rules, syllable structure) are pluggable modules. We present a detailed adaptation framework for Turkish, Finnish, Korean, and other agglutinative languages.

*Keywords:* morphological tokenization, agglutinative languages, Tamil, Dravidian, Turkic, Uralic, Korean, subword segmentation, fertility, tokenizer architecture

# 1. Introduction

The success of transformer-based language models has been built on a tokenization assumption: that statistically-derived subword vocabularies, constructed by BPE (Sennrich et al., 2016) or Unigram (Kudo, 2018) algorithms, provide a sufficient encoding of natural language text. For English—with limited morphological complexity and clear word boundaries—this assumption holds. A BPE tokenizer achieves approximately 1.2–1.4 tokens per word on English text.

However, roughly 30–40% of the world's languages are *agglutinative*: they build words by chaining morphemes onto a root. This typological class spans multiple continents and language families, affecting over one billion speakers. A Turkish word like *anlayabildiklerimizden* ("from the things we were able to understand") contains six morphemes: anla- (understand) + -yabil (ability) + -dik (nominalizer) + -ler (plural) + -imiz (1pl possessive) + -den (ablative). A Tamil word like படித்துக்கொண்டிருக்கிறேன் ("I am studying") contains five morphemes fused into one orthographic unit. A Finnish word like *talossanikinko* ("in my house, too?") chains talo + -ssa + -ni + -kin + -ko. BPE, lacking any knowledge of morpheme boundaries, fragments such words into statistically convenient but linguistically meaningless substrings.

This paper presents VerChol, a language-parametric tokenizer architecture built from grammatical rules rather than corpus statistics. Our goal is to investigate whether grammar-first tokenization produces fundamentally more efficient encodings for agglutinative languages, and to provide a concrete framework for adapting the architecture across language families. We implement and evaluate on Tamil as a representative case, validate on the full Tamil Wikipedia corpus, and present detailed adaptation requirements for Turkish, Finnish, Korean, and the broader Dravidian family.

# 2. The Agglutinative Tokenization Problem

## 2.1. Scope of Agglutinative Languages

Agglutinative morphology—where grammatical relationships are expressed by concatenating discrete morphemes—is one of the most common typological patterns worldwide:

| Family | Major Languages | Speakers | Morphological Pattern |
|---|---|---|---|
| **Dravidian** | Tamil, Kannada, Telugu, Malayalam | ~280M | Root + case + PNG + postposition |
| **Turkic** | Turkish, Azerbaijani, Uzbek, Kazakh | ~200M | Root + voice + tense + PNG |
| **Uralic** | Finnish, Hungarian, Estonian | ~25M | Root + case (15+) + possessive |
| **Koreanic** | Korean | ~80M | Root + honorific + tense + mood |

| Japonic | Japanese | ~125M | Root + polite + tense + aspect |
| Bantu | Swahili, Zulu, Xhosa, Shona | ~310M | Prefix + root + suffix chains |
| Isolate | Basque | ~1M | Root + case + auxiliary chains |
| Mongolic | Mongolian, Buryat | ~10M | Root + case + possessive + mood |
| Quechuan | Quechua, Aymara | ~10M | Root + benefactive + evidential |

*Table 1. Major agglutinative language families. Over one billion speakers worldwide use languages where statistical tokenizers systematically underperform.*

These languages share a critical structural property: a finite set of roots combines with a finite set of suffixes (or prefixes, in Bantu) through productive rules to generate an effectively infinite set of surface word forms. This combinatorial productivity is precisely what BPE cannot exploit—it has no concept of "root" or "suffix," only byte frequencies.

## 2.2. The Token Efficiency Penalty

The token efficiency penalty for agglutinative languages is well-documented. Multilingual BPE tokenizers produce fertility rates of 2–16 tokens per word for agglutinative languages, compared to approximately 1.2–1.4 for English. Even purpose-built tokenizers show significant penalties:

| Language | Family | Best BPE* | Llama-3.1 | GPT-4o | Gemma-2 |
|---|---|---|---|---|---|
| English | Germanic | 1.24 | 1.24 | 1.23 | 1.23 |
| Hindi | Indo-Aryan | 1.40 | 2.67 | 1.65 | 1.96 |
| **Tamil** | Dravidian | **2.17** | 12.39 | 3.17 | 4.19 |
| Kannada | Dravidian | 2.37 | 14.95 | 3.29 | 5.55 |
| Telugu | Dravidian | 2.14 | 13.30 | 3.06 | 4.57 |
| Malayalam | Dravidian | 2.85 | 16.26 | 3.52 | 5.88 |
| Turkish | Turkic | ~1.8 | ~2.5 | ~1.8 | ~2.2 |
| Finnish | Uralic | ~2.0 | ~3.0 | ~2.0 | ~2.4 |
| Korean | Koreanic | ~1.5 | ~2.3 | ~1.5 | ~1.9 |
| Swahili | Bantu | ~1.8 | ~2.8 | ~1.8 | ~2.3 |

*Table 2. Fertility on curated benchmarks. *Best published language-specific or multilingual BPE. Dravidian: Sarvam-1 68K. Turkish/Finnish/Korean/Swahili: estimated from Occiglot (2024), Toraman et al. (2022), and survey literature.*

The Dravidian languages cluster at 2.1–2.85 even with a dedicated 68K Indic tokenizer, while Uralic and Turkic languages show similar penalties. Malayalam (the most agglutinative Dravidian language) and Finnish (with 15 grammatical cases) are consistently the hardest for BPE. This correlation between morphological complexity and tokenization difficulty is structural: BPE cannot model what it cannot represent. Research from independent language communities converges on the same conclusion: Toraman et al. (2022) show morphology-aware tokenization recovers 97% of the performance of models 3x larger for Turkish; Thunder-Tok (Kim et al., 2025) achieves 10% fertility reduction for Korean; and Creutz and Lagus (2007) demonstrate Morfessor's advantage for Finnish. *Grammar is a more efficient tokenization prior than statistics for agglutinative languages.*

### 2.3. Why BPE Fails: Three Structural Failure Modes

**Boundary violation.** BPE merges bytes across morpheme boundaries when the resulting pair is statistically frequent. Tamil வீடுகளுக்கு ("for the houses" = வீடு + கள் + க்கு) gets split as வீட|ுகளுக்கு. Turkish evlerinden ("from their houses" = ev + ler + in + den) may split as evle|rinden. Finnish taloissaan ("in their houses" = talo + i + ssa + an) may split as talois|saan. In every case, the morpheme boundary is invisible to the algorithm.

**Root opacity.** Fragmented roots prevent the model from learning shared structure. Tamil வீடு/வீட்டை/வீடுகள், Turkish ev/evi/evden/evlerin, Finnish talo/talossa/taloista—these share roots, but BPE makes each an opaque subword sequence with no shared structure.

**Long-tail fragmentation.** Agglutinative languages have extremely long tails: Tamil Wikipedia has 1.85M unique types (60.6% hapax); Turkish Wikipedia has comparable ratios. BPE has no generalization mechanism for unseen morphological combinations—it can only fragment them into smaller pieces.

### 2.4. Related Work

Morphology-aware tokenization has been explored across several agglutinative languages independently, but always as language-specific solutions. Morfessor (Creutz and Lagus, 2007) applies MDL-based unsupervised morphological segmentation to Finnish. Toraman et al. (2022) analyze tokenization impact on Turkish language modeling. Thunder-Tok (Kim et al., 2025) achieves 10% fertility reduction for Korean through linguistically-informed pre-tokenization. The TurkishTokenizer (Yılmaz et al., 2025) combines morphological analysis with BPE fallback and a phonological restoration mechanism. MorphBPE (2025) extends BPE with morphology-aware constraints across multiple languages. For Indian languages, Sarvam AI trained a 68K BPE tokenizer on 2 trillion Indic tokens, and the miLLi project builds a root-dictionary-based tokenizer for Azerbaijani. Our contribution is to unify these language-specific insights into a single language-parametric architecture where the pipeline logic is invariant and only the linguistic modules need replacement.

## 3. VerChol Architecture

VerChol implements a four-tier tokenization pipeline where each tier provides progressively coarser segmentation. The key design principle is that no word is ever fragmented into linguistically invalid units: every token is a morpheme, syllable, or character. The architecture is language-parametric: the tier logic and scoring are invariant; only four pluggable modules change per language.

| Tier | Strategy | Token Type | Language Module | Tokens/Word |
|---|---|---|---|---|
| 0 | Whole-word vocab lookup | Single token | Vocabulary file | 1 |
| 1 | Morphological decomposition | Root + suffix(es) | Dictionary + suffix rules | 2–4 |
| 2 | Syllable segmentation | CV/CVC syllables | Phonotactic rules | 3–8 |
| 3 | Character fallback | Individual chars | Script character table | 5–15 |

Table 3. The four-tier pipeline. The "Language Module" column shows which component is swapped for each new language.

### 3.1. Tier 0: Whole-Word Vocabulary

The vocabulary includes morphologically-generated inflected forms validated against a reference corpus. For Tamil (32,991 tokens), **35.5% of 483,313 unique Wikipedia words matched at Tier 0** as single tokens. The vocabulary is constructed—not trained—from linguistic knowledge in three phases:

| Phase | Tamil (actual) | Strategy (language-invariant) |
|---|---|---|
| 1. Base linguistic | ~13,000 | Root dictionary + suffix catalog + script syllables + special characters |
| 2. Morphological generation | ~10,000 | Root × suffix combinations validated against corpus (≥3 occurrences) |
| 3. Corpus high-frequency | ~10,000 | Attested whole words from corpus not covered by Phases 1–2 |
| **Total** | **~32,991** | Every token is a morpheme, valid inflection, or structural unit |

Table 4. Vocabulary construction phases. This three-phase strategy transfers directly to any agglutinative language.

### 3.2. Tier 1: Morphological Decomposition

For words not in the vocabulary, VerChol applies rule-based morphological analysis. The Tamil implementation maintains 11,974 root forms with oblique stems and 266 grammatical suffixes (8 cases, tense markers, PNG agreement, verbal participles, postpositions). Decomposition is surface-aligned: the word is split into contiguous spans that concatenate back to the original text, guaranteeing 100% roundtrip fidelity. This tier handles the majority of the vocabulary: 55.5% of all Tamil Wikipedia words are resolved by morphological decomposition into 2–3 tokens each.

### 3.3. Verb Chain Decomposition

Agglutinative languages frequently chain auxiliary verbs with tense, aspect, and mood markers. Tamil chains root + verbal participle + progressive auxiliary + tense + PNG agreement; Turkish chains root + voice + tense + person. VerChol's Tamil implementation recognizes 12 auxiliary verb patterns. The architecture supports registering language-specific auxiliary patterns as decomposition rules.

### 3.4. Tier 2–3: Syllable and Character Fallback

Words that cannot be decomposed morphologically are segmented into phonologically-valid syllables (Tier 2) or individual characters (Tier 3). Syllable rules are language-specific (Tamil CV/CVC patterns differ from Finnish CVCC or Korean C–V–C jamo blocks) but the mechanism is universal. Only 9% of Tamil Wikipedia words require this fallback.

## 4. Experimental Evaluation on Tamil

### 4.1. Corpus and Benchmark Design

We evaluate on the **full Tamil Wikipedia corpus**: 774 MB containing 30,499,574 word occurrences and 1,849,306 unique Tamil word types. We construct two evaluation sets:

**Stratified benchmark (3,000 words):** 1,000 most common + 1,000 medium-frequency + 1,000 rare words (frequency 2–5). Average word length: 9.87 characters.

**Full evaluation (483,313 words):** All unique Tamil words appearing 3 or more times in the corpus. Average word length: 10.54 characters.

We compare four tokenizers: (1) VerChol 32K (morphological, 32,991 vocab); (2) VerChol 16K (base morphological, 12,991 vocab); (3) SentencePiece BPE (trained on same corpus, 16K vocab); (4) Sarvam-1 (production Indic BPE, 68K vocab, from HuggingFace).

### 4.2. Results: Stratified Benchmark

| Method | Vocab | Words | Tokens | Fertility | vs BPE 16K |
|---|---|---|---|---|---|
| **VerChol 32K (morphological)** | 32,991 | 3,000 | **4,828** | **1.61** | +33.1% |
| VerChol 16K (morphological) | 12,991 | 3,000 | 5,234 | 1.74 | +27.5% |
| SentencePiece BPE 16K | 16,000 | 3,000 | 7,215 | 2.41 | baseline |
| Sarvam-1 (68K Indic BPE) | 68,096 | 3,000 | 9,088 | 3.03 | −26.0% |

Table 5. Stratified benchmark (3,000 words: common + medium + rare). "vs BPE 16K" shows token reduction (+) or inflation (−).

## 4.3. Results: Full Wikipedia Evaluation

| Method | Vocab | Words | Tokens | Fertility | vs BPE 16K |
|---|---|---|---|---|---|
| **VerChol 32K (morphological)** | 32,991 | 483,313 | **896,595** | **1.86** | +34.8% |
| VerChol 16K (morphological) | 12,991 | 483,313 | 914,619 | 1.89 | +33.5% |
| SentencePiece BPE 16K | 16,000 | 483,313 | 1,375,778 | 2.85 | baseline |
| Sarvam-1 (68K Indic BPE) | 68,096 | 483,313 | 1,699,970 | 3.52 | −23.6% |

Table 6. Full Tamil Wikipedia evaluation (483,313 unique words, freq ≥ 3). This is the primary result.

VerChol 32K achieves **1.86 fertility on the full evaluation**, producing 35% fewer tokens than BPE and 47% fewer than the Indic-optimized BPE, with a vocabulary half the size (33K vs 68K). The base VerChol 16K achieves nearly identical 1.89, demonstrating that the morphological engine—not vocabulary size—is the primary compression mechanism.

## 4.4. Tier Distribution

| Tier | Stratified (3K) | | Full (483K) | |
|---|---|---|---|---|
| Tier 0: Whole-word lookup | 1,643 | 54.8% | 171,645 | 35.5% |
| Tier 1: Morphological decomp. | 1,161 | 38.7% | 268,368 | 55.5% |
| Tier 2+: Syllable/char fallback | 196 | 6.5% | 43,300 | 9.0% |

Table 7. VerChol 32K tier distribution. 91% of words resolved by linguistic analysis (Tiers 0+1).

The tier distribution reveals the architecture's generalization power. On common words, Tier 0 dominates (54.8%). On the full long-tail vocabulary, Tier 1 becomes the workhorse (55.5%), correctly decomposing hundreds of thousands of rare compound and inflected forms. Only **9% of words require fallback**—primarily proper nouns, transliterations, and technical terms.

## 4.5. Curated Benchmarks vs. Real-World Fertility

The Sarvam-1 tokenizer shows 3.52 fertility here vs. its published 2.17 on Flores benchmarks. This reflects evaluation methodology, not tokenizer quality. Curated benchmarks consist of common, well-formed sentences. Wikipedia's 483K unique words include rare compounds, agglutinated technical terms, and inflected proper nouns—exposing the fundamental limitation of statistical tokenizers on the long tail. VerChol's morphological pipeline *generalizes by construction*: any known root with known suffixes is decomposed correctly, regardless of whether that combination appeared in training data. This generalization advantage should be even more pronounced for Turkish (productive vowel harmony) and Finnish (15 cases generating enormous surface form inventories).

## 5. Adaptation Framework for Other Languages

The VerChol architecture is designed as a language-parametric framework. The four-tier pipeline, surface-aligned splitting, vocabulary construction phases, and scoring logic transfer directly to any agglutinative language. Only four language-specific modules need implementation:

| Component | Tamil (done) | Turkish | Finnish | Korean | Swahili |
|---|---|---|---|---|---|
| **Root dict.** | 11,974 roots | TDK dictionary | Kotus word list | Korean morph. DB | TUKI dictionary |
| **Suffix catalog** | 266 (8 cases) | ~100 (6 cases) | ~200 (15 cases) | ~150 (particles) | ~80 (prefixes) |
| **Phonology** | Sandhi changes | Vowel harmony | Consonant grad. | Jamo composition | Nasal harmony |
| **Syllables** | CV/CVC Tamil | CV/CVC/CVCC | CV/CVC/CVCC | C–V–C jamo | CV/CVC Bantu |
| **Verb chains** | 12 aux patterns | Voice+mood | Passive+potential | Honorific chain | Tense+aspect prefix |
| **Script** | Tamil Brahmic | Latin (Tr) | Latin (Fi) | Hangul featural | Latin (Sw) |

Table 8. Language-specific modules for five agglutinative languages. Pipeline logic and tier architecture are invariant.

## 5.1. Dravidian Family: Direct Transfer

Kannada (published fertility 2.37), Telugu (2.14), and Malayalam (2.85) share Tamil's morphological architecture: root + case suffix chain with morphophonemic changes at boundaries. The suffix inventory differs (Kannada 8 cases, Telugu 7, Malayalam 8) but decomposition logic is identical. We estimate a Kannada or Telugu implementation would require approximately **one week of linguistic engineering** using existing Dravidian grammar resources.

## 5.2. Turkish: Vowel Harmony Adaptation

Turkish presents a clean case for VerChol adaptation. Its agglutinative morphology is well-studied (Türk Dil Kurumu provides comprehensive root lists and suffix catalogs). The primary adaptation challenge is vowel harmony: Turkish suffixes have 2–4 allomorphic variants depending on the final vowel of the stem (-ler/-lar, -de/-da/-te/-ta). VerChol's surface-aligned splitting handles this naturally—the suffix catalog includes all allomorphic variants, and the decomposer matches the surface form rather than an underlying abstract morpheme. The TurkishTokenizer (Yılmaz et al., 2025) has demonstrated the viability of this approach with its morphology+BPE hybrid achieving the best downstream performance across Turkish benchmarks.

## 5.3. Finnish: Deep Case System

Finnish is a stringent test case: 15 grammatical cases, productive compounding, and consonant gradation (pata/padan, kukka/kukan). The Morfessor line of work (Creutz and Lagus, 2007) has established that morphological segmentation improves Finnish NLP. VerChol's tiered approach would place the Kotus word list (~90,000 entries) and the complete case suffix table (~200 forms including possessive and clitic combinations) into the linguistic base, with consonant gradation patterns encoded as phonological rules in Tier 1. The large number of cases generates enormous surface form inventories, making the long-tail generalization advantage of morphological decomposition particularly valuable.

### 5.4. Korean: Jamo-Level Morphology

Korean presents a unique adaptation case because Hangul is a featural alphabetic syllabary: each syllable block encodes onset (initial consonant), nucleus (vowel), and optional coda (final consonant) as jamo. Thunder-Tok (Kim et al., 2025) has demonstrated that aligning tokenization with jamo structure yields 10% fertility improvement. VerChol would decompose Korean agglutination at the morpheme level (root + particles + verbal endings) while preserving syllable block integrity at Tier 2, combining the morphological advantage with jamo-aware syllabification.

### 5.5. Bantu Languages: Prefix+Suffix Chains

Bantu languages (Swahili, Zulu, Xhosa) present a mirror-image challenge: they are agglutinative but use prefix chains as well as suffixes. Swahili ninawapenda ("I love them" = ni-na-wa-pend-a) chains subject + tense + object + root + mood. VerChol's Tier 1 decomposer would need to be extended to handle left-branching (prefix) morphology in addition to right-branching (suffix). The four-tier structure and vocabulary construction strategy remain applicable.

### 5.6. Estimated Effort and Resources

For each new language, the implementation requires: (a) a root dictionary (available from national language institutes or open lexical resources like Wiktionary), (b) a suffix/prefix catalog from grammatical references, (c) sandhi/phonological rules from linguistic literature, and (d) a reference corpus for vocabulary Phase 2–3 validation (Wikipedia dumps are freely available for all target languages). No GPU training, no trillion-token corpora, no iterative merge algorithms. The vocabulary is constructed in minutes on a standard computer.

## 6. Discussion

### 6.1. Grammar vs. Statistics at Scale

The core finding is an efficiency asymmetry. The Indic BPE baseline required 1,024 H100 GPUs for 5 days, training on 2 trillion tokens, to produce a 68K vocabulary achieving 3.52 fertility on our full evaluation. VerChol achieves 1.86 with half the vocabulary and zero training compute.

This 47% token reduction was achieved from a dictionary and grammatical rules alone. For agglutinative languages, linguistic knowledge is a fundamentally more efficient compression prior than statistical scale.

### 6.2. The Long Tail Effect

Statistical tokenizers degrade on the long tail; morphological tokenizers do not. VerChol's fertility moves from 1.61 (common words) to 1.86 (full corpus)—a modest 15% degradation. The Indic BPE moves from 3.03 to 3.52—a 16% degradation on an already high base. VerChol's Tier 1 *generalizes by construction*: any combination of known root + known suffix is decomposed correctly, even if that combination never appeared in any corpus. This property is especially valuable for languages with extreme combinatorial productivity (Finnish with 15 cases, Turkish with vowel harmony variants).

### 6.3. Implications for Model Training

A fertility reduction from 3.52 to 1.86 cascades through the entire LLM pipeline: 47% more text per context window (effectively doubling Tamil content per pass), stronger learning signal from morphologically valid tokens (the model learns that the dative suffix -க்கு has the same function across thousands of words), and a smaller embedding table (33K vs 68K). For resource-constrained language modeling—particularly relevant for Basque, Estonian, Mongolian, Quechua, and other agglutinative languages that lack trillion-token corpora—morphological tokenization extracts maximum learning signal from limited data.

### 6.4. Toward a Universal Morphological Tokenizer

The convergence of independent language-specific efforts—Morfessor for Finnish, TurkishTokenizer, Thunder-Tok for Korean, miLLi for Azerbaijani, VerChol for Tamil—points toward a shared architectural insight: agglutinative languages benefit from grammar-aware tokenization. VerChol's contribution is to formalize this as a language-parametric framework where the pipeline is invariant and only linguistic modules are swapped. A natural next step is a meta-framework that generates morphological analyzers from machine-readable grammars (such as those in the Universal Dependencies or Apertium projects), reducing per-language effort from weeks to hours.

### 6.5. Limitations

We have validated the architecture on only one language (Tamil); implementations for Turkish, Finnish, and Korean are planned. Root preservation accuracy (58%) has room for improvement. We have not evaluated downstream task performance. The architecture assumes whitespace-delimited text and would need adaptation for Chinese, Thai, or unsegmented Japanese. The 9% fallback rate could be reduced by expanding the root dictionary. Speed is a trade-off: morphological analysis is slower than BPE lookup (1,305 vs 96,663 words/sec), though this affects offline tokenization, not model inference.

## 7. Conclusion

We have presented VerChol, a language-parametric morphological tokenizer architecture validated on the full Tamil Wikipedia corpus (774 MB, 483,313 unique words). VerChol achieves **1.86 fertility**—a 35% reduction over standard BPE and 47% over a production Indic-optimized tokenizer—using a vocabulary half the size and zero training compute. 91% of all unique Wikipedia words are handled by linguistic analysis.

The architecture is designed for adaptation to all agglutinative languages: Turkish, Finnish, Korean, Swahili, Kannada, Telugu, Malayalam, Hungarian, Basque, Mongolian, and others. The pipeline logic—four tiers, surface-aligned decomposition, three-phase vocabulary construction—is invariant. Only the root dictionary, suffix catalog, phonological rules, and syllable structure change per language. The key insight is that agglutinative morphology—the defining feature of over a billion speakers' languages—is itself an optimal encoding scheme, and a tokenizer that understands morphological structure will always outperform one that merely counts byte frequencies, regardless of training scale.

The path forward for agglutinative languages worldwide is not to train larger BPE models on more data, but to encode the linguistic knowledge that billions of speakers already possess. As Tolkappiyam—the old Tamil grammar—recognized long before: the structure of language is not discovered by statistics; it is known by its speakers.

## Appendix A: Tamil Wikipedia Corpus Statistics

| Metric | Value |
|---|---|
| Corpus | Tamil Wikipedia dump |
| Corpus size | 774 MB |
| Total word occurrences | 30,499,574 |
| Unique Tamil words | 1,849,306 |
| Words with freq ≥ 100 | 28,768 |
| Words with freq 10–99 | 147,773 |
| Words with freq 5–9 | 116,758 |
| Words with freq 3–4 | 174,845 |
| Words with freq 2 | 244,723 |
| Hapax legomena (freq 1) | 1,121,270 (60.6%) |
| Evaluation set (freq ≥ 3) | 483,313 words |
| Avg word length (eval set) | 10.54 characters |

Table A1. Tamil Wikipedia corpus statistics. The 60.6% hapax rate reflects productive agglutinative morphology.